\documentclass[runningheads]{llncs}

\usepackage{eccvabbrv}

\usepackage{graphicx}
\usepackage{booktabs}
\usepackage{multirow}
\usepackage{colortbl}
\usepackage{wrapfig}
\usepackage{xcolor}
\usepackage{subcaption}
\usepackage[accsupp]{axessibility}  

\usepackage{hyperref}

\usepackage{orcidlink}

\begin{document}

\title{Inclusive Interactive Collisions for Multi-View Consistent Compositional 3D Generation} 

\titlerunning{I\textsuperscript{2}C-3D}

\author{Chang Liu\inst{1}\and
Mingwen Shao\inst{1,2}\and
Xiang Lv\inst{1}\and
Xinyuan Chen\inst{1} \and
Lingzhuang Meng\inst{1} \and
Qiao Zhang\inst{1} \and
Zhengyi Gong\inst{1} \and
Jinghao Hu\inst{3}}

\authorrunning{C.~Liu et al.}

\institute{School of Computer Science and Technology, China University of Petroleum (East China), Qingdao, 266580, China \and
Artificial Intelligence Research Institute, Shenzhen University of Advanced Technology, Shenzhen, 518107, China \and
College of Computer Science, Northwest University, Xi'an, 710127, China \\}

\maketitle

\begin{abstract}
Recent breakthroughs in 3D generation have advanced notably with the development of text-to-image diffusion model. However, existing methods remain two practical challenges: (1) They primarily generate single 3D object, but struggle to generate multi-object compositional 3D assets due to the lack of the modeling for Gaussian primitives in reasonable interactions. (2) They often suffer from cross-view inconsistency during 3D optimization, as Score Distillation Sampling inherently performs on each single view, inevitably resulting in cross-view hallucinations. To solve above issues, we propose I\textsuperscript{2}C-3D, a novel optimization-based method to generate multi-view consistent compositional 3D assets with reasonable interactions. Specifically, we propose an Inclusive Interactive Collisions strategy to guide Gaussian primitives appearing in reasonable interaction regions naturally, thereby ensuring objects in the compositional scene interact in a physically plausible and visually coherent way. Additionally, to enhance multi-view consistency, Multi-View Adaptive Score Distillation Sampling is devised to distill multi-view consistency prior and layout prior from pre-trained diffusion model by modulating attention map of instance token and spatial token across viewpoints. Benefiting from above elaborate designs, I\textsuperscript{2}C-3D not only generates high-fidelity multi-view consistent compositional 3D assets but also supports 3D editing flexibly, facilitating complex scene generation. Extensive experiments demonstrate our I\textsuperscript{2}C-3D outperforms existing methods in generation quality and multi-view consistency.
\end{abstract}

\section{Introduction}
\label{Introduction}

Recently, generating high-quality 3D assets according to given text prompt or a single image is a growing demand in AR/VR \cite{ar,vr}, movies and industrial design \cite{gongye}. Thanks to large scale text-to-image generative diffusion models \cite{ddpm,ddim,SD}, significant progress has also been made in the field of 3D generation \cite{cite1,cite2}. Previous works have been broadly divided into two categories: (1) Training-based generation \cite{instant3d,lrm,lgm}, which is a feed-forward generation process by training an efficient reconstruction model on 3D data. (2) Optimization-based generation \cite{lnerf,magic3d,sjc}, which distillates prior knowledge from pre-trained text-to-image diffusion model via Score Distillation Sampling (SDS) \cite{sds} for 3D generation \cite{gsgen,chen2024vp3d,prolificdreamer}.\par

\begin{figure*}[t]
  \centering
  \begin{subfigure}{0.2\linewidth}
    \includegraphics[width=\linewidth]{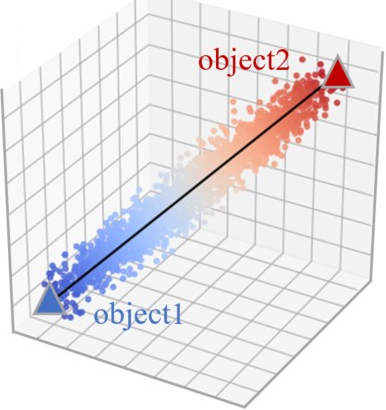}   
    \caption{Visualization.}
    \label{fig:short-a}
  \end{subfigure}
  \hfill
  \begin{subfigure}{0.77\linewidth}
    \includegraphics[width=\linewidth]{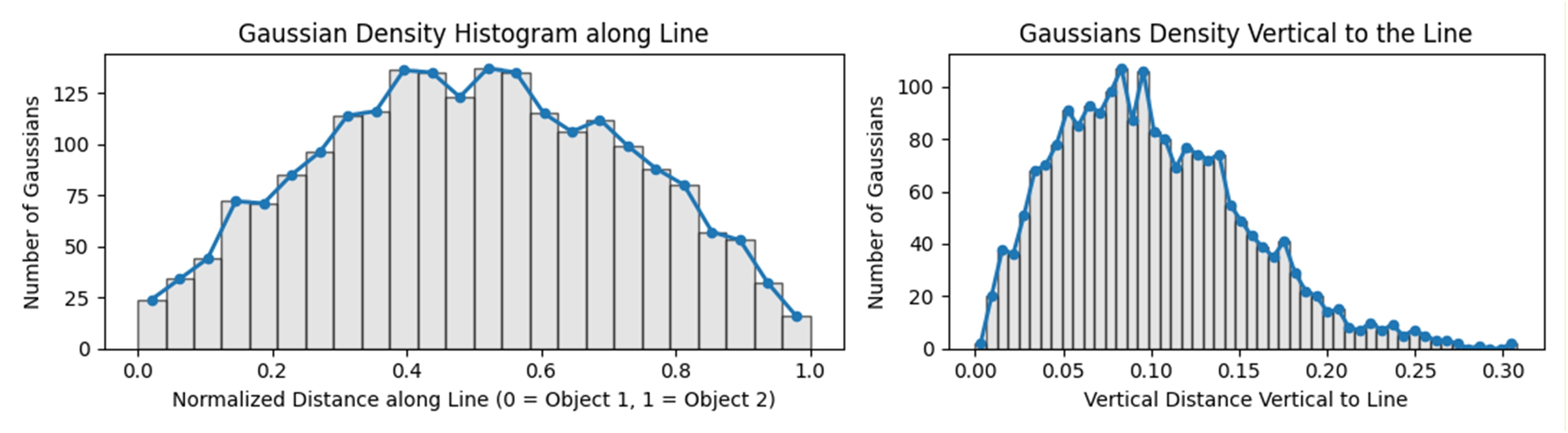}   
    \caption{Analysis of the distribution of Gaussian primitives.}
    \label{fig:short-b}
  \end{subfigure}
  \caption{\textbf{Analysis and visualization of the statistical distribution of Gaussian primitives in the interaction region.} (a) The Gaussian primitives are sparse at the line’s ends and densely distributed with more collision near the midpoint of the line. (b) Through quantifying the Gaussian distribution along line and vertical line, further demonstrate most Gaussian primitives are distributed around the midpoint of line.}
  \label{fig:short}
\end{figure*}


Although the above methods can achieve remarkable results in single-object 3D generation, they struggle to generate multi-object scene with complex interactions. Specifically, for training-based methods, they lack complex multi-object data in training dataset \cite{objaverse}, which inevitably leads to failure on multi-object 3D generation. For optimization-based methods, they struggle to distill compositional prior knowledge from pre-trained text-to-image diffusion models, which is mainly due to the fact that existing pre-trained 2D diffusion models lack fine-grained controllability, further fail to precisely follow the text prompt involving multiple objects, attributes or spatial relationships \cite{ToMe,boxdiff,gligen}.\par

We thoroughly investigate the statistical distribution of Gaussian primitives in the interaction region and observe that Gaussian primitives tend to concentrate around the midpoint of the connecting line between interacting objects, forming a roughly cylindrical interaction space. Specifically, Fig. \ref{fig:short-a} visualizes the distribution of Gaussian primitives around the connecting line of the two interacting objects: blue and red dots denote Gaussian primitives belonging to objects 1 and object 2, respectively. Lighter-colored regions indicate that the two kinds of Gaussian primitives interact and collide. The results in Fig. \ref{fig:short-a} show the Gaussian primitives at the ends of the connecting line are relatively sparse, and those located near the midpoint of the line are densely distributed with more collision. Moreover, we plot the Gaussian Density Histogram along Line and Gaussians Density Vertical to the Line (displayed in Fig. \ref{fig:short-b}), which quantitatively indicate the Gaussian density near the midpoint of the line is larger. Inspired by these observations, we design a collision constraint that guides Gaussian primitives near the interaction boundary to a reasonable collision region, enabling realistic interaction-aware 3D generation.\par

\begin{figure*}[t]
\centering
\includegraphics[width=1.0\textwidth]{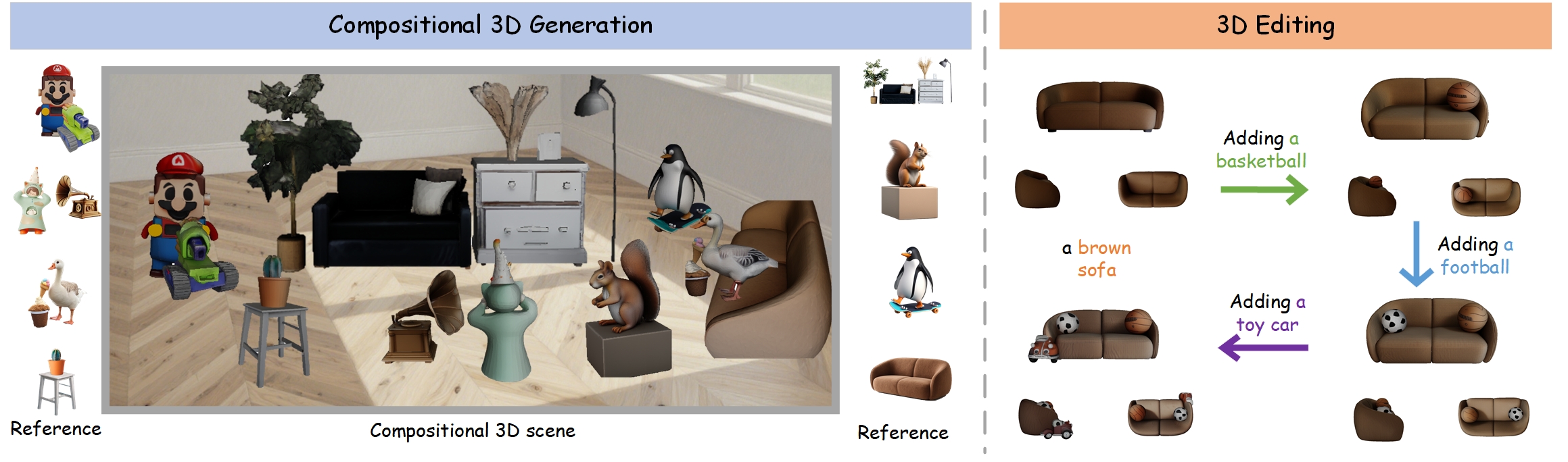} 
\caption{\textbf{Illustration of our I\textsuperscript{2}C-3D generated compositional 3D scene and 3D editing results.} Our I\textsuperscript{2}C-3D not only generates multi-view consistent 3D scene with reasonable interaction region, but also achieves flexible 3D editing.}
\label{fig1}
\end{figure*}

In this paper, we propose the I\textsuperscript{2}C-3D, a novel framework to generate multi-view consistent compositional 3D assets with reasonable interactions, which is a coarse-to-fine optimization process. In coarse 3D generation stage, we elaborate a local-global 3D layout generator based on LLMs, then utilize pre-trained single-object 3D generation model to reconstruct and compose a coarse 3D scene. In refinement stage, we propose Inclusive Interactive Collisions strategy, which not only constrains most Gaussian primitives are in inside the bounding box to ensure generated compositional 3D scene aligns with the planned 3D layout, but also guide Gaussian primitives to collide naturally in interaction region for reasonable interaction generation. Subsequently, Multi-View Adaptive Score Distillation Sampling is designed to distill multi-view consistency prior and fine-grained layout prior from pre-trained diffusion model through adaptive attention modulation for enhancing multi-view consistency. Attribute to above elaborate designs, our I\textsuperscript{2}C-3D not only generates multi-view consistent compositional 3D scene with reasonable interaction, but also supports flexible 3D editing (see Fig. \ref{fig1}). Extensive quantitative and qualitative experiments suggest our I\textsuperscript{2}C-3D achieves state-of-the-art (SOTA) results, outperforming in high quality and multi-view consistency. Our main contributions are summarized as follows:

\begin{itemize}
\item We propose I\textsuperscript{2}C-3D, a novel optimization-based compositional 3D generation framework, which can generate multi-view consistent compositional 3D assets with reasonable interactions.
\item We devise an Inclusive Interactive Collisions (I\textsuperscript{2}C) strategy to constrain gaussian primitives appearing in reasonable interaction regions, realizing physically plausible interaction.
\item Multi-View Adaptive Score Distillation Sampling (MV-ASDS) is designed to distill cross-view prior and fine-grained layout prior for achieving multi-view consistency.
\item Extensive experiments show our I\textsuperscript{2}C-3D qualitatively and quantitatively outperforms SOTA methods.
\end{itemize}

\section{Related Work}

\subsection{Text-to-3D Generation}
Significant advances have been made in text-to-3D generation with the development of generative diffusion models. A representative work, DreamFusion \cite{poole2022dreamfusion} creatively proposes Score Distillation Sampling to distill the prior from powerful 2D diffusion models for 3D generation. Inspired by DreamFusion, various excellent approaches have been proposed. On the one hand, researchers attempt to explore more efficient 3D representations, from NeRF \cite{nerf} to 3D Gaussians (3DGS) \cite{3dgs,wan2025s2gaussian,wan2026splatweaver}, DMTet \cite{dmtet} and hybrid representations \cite{SceneWiz3D}. On the other hand, they exploit more efficient supervision to improve over-saturation and over-smoothing in SDS, such as VSD \cite{prolificdreamer} and CSD \cite{CSD}. Specifically, Magic3D \cite{magic3d} devises a coarse-to-fine optimization strategy that utilizes multiple diffusion priors at different resolutions to optimize the 3D representation for high-resolution 3D Generation. Subsequently, GaussianDreamer \cite{gaussiandreamer} bridges 2D and 3D diffusion models for fast text-to-3D Generation. Meanwhile, gsgen \cite{gsgen} proposes a two-stage optimization strategy that first joints 2D and 3D diffusion guidance to shape a rough geometric structure, then refines appearance details through compactness-based densification. However, these methods can only generate single-object 3D asset, but our I\textsuperscript{2}C-3D supports multi-object 3D generation.\par
\subsection{Compositional 3D generation}
Compositional 3D generation aims to generate attribute-specific and spatially reasonable multi-object compositional 3D assets from text prompt or an image. Specifically, ComboVerse \cite{comboverse} proposes spatially-aware score distillation sampling (SSDS) to guide the position of objects for multi-object combination. Meanwhile, REPARO \cite{reparo} designs differentiable rendering techniques to optimize the layout of meshes, ensuring coherent scene composition. GALA3D \cite{zhou2024gala3d} utilizes layout priors to bridge text and compositional scene for end-to-end high-quality scene-level 3D content generation. Subsequently, CompGS \cite{ge2025compgs} employs 3DGS initialization with 2D compositionality and Dynamic SDS Optimization for high-quality 3D generation. Layout-your-3D \cite{layout-your-3d} exploits 2D layout as blueprints, enabling precise and plausible control over 3D generation. However, above methods struggle to generate multi-view consistent compositional 3D assets with plausible interaction. In contrast, our I\textsuperscript{2}C-3D constrains the distribution of Gaussian primitives in interaction region based on geometry theory for natural interaction generation. \par

\section{Methodology}
\subsection{Preliminary}
\textbf{Score Distillation Sampling. }DreamFusion first proposes Score Distillation Sampling (SDS) for distilling prior from pre-trained 2D diffusion models to optimize 3D representation. Specifically, given a text prompt $y$ and an initialized 3D representation parameterized by $\theta$, the rendered image $x$ from viewpoint $c$ can be obtained through a differentiable rendering process $g(\theta,c)$. Subsequently, add Gaussian noise $\epsilon$ to $x$ and leverage a pre-trained 2D diffusion model $\phi$ to predict noise, then denoise it. By computing the difference between predicted noise $\epsilon_{\phi }(x_{t},y,t)$ and initial noise $\epsilon$, SDS constructs gradient and propagates backward it to optimize 3D representation, which can be expressed as:
\begin{eqnarray}
\bigtriangledown _{\theta }\mathcal{L}_{SDS} = \mathbb{E} _{\epsilon ,t}\left [w(t)(\epsilon_{\phi }(x_{t},y,t)-\epsilon)\frac{\partial g(\theta,c)}{\partial \theta } )  \right ] ,
\end{eqnarray}
where $t$ is timestep, $w(t)$ is a weighting function. DreamFusion utilizes NeRF \cite{nerf} as 3D representation and renders the image via voxel rendering. While considering efficiency and editability, we choose 3DGS as the 3D representation for subsequent optimization. \par
\begin{figure*}[t]
\centering
\includegraphics[width=1.0\textwidth]{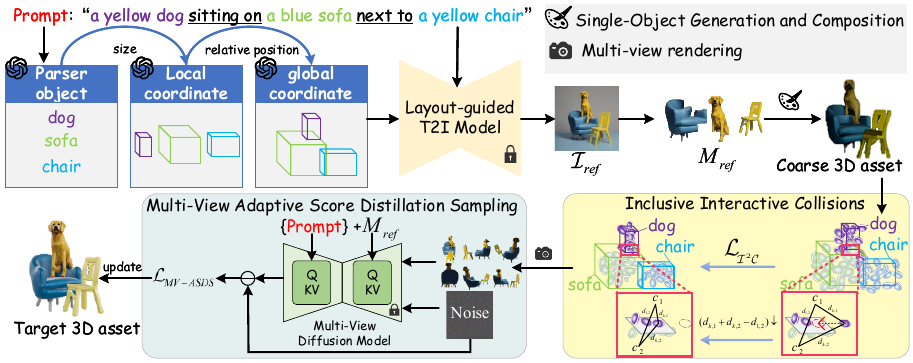} 
\caption{\textbf{Overview of our I\textsuperscript{2}C-3D. }We first utilize pre-trained single-object 3D generation model to reconstruct and compose a coarse 3D scene. Subsequently, leverage Inclusive Interactive Collisions strategy (I\textsuperscript{2}C) to refine interaction region generation. Then Multi-View Adaptive Score Distillation Sampling (MV-ASDS) is applied to distill cross-view priors from multi-view diffusion model for achieving multi-view consistency. Through coarse-to-fine optimization, high-quality 3D asset has been generated.}
\label{fig3}
\end{figure*}


\subsection{Coarse 3D Generation Stage}
\textbf{3D layout generation. }Unlike 2D layout generation, LLMs struggle to directly derive 3D coordinates from complex descriptions because their training data lack sufficient 3D-text pairs. Consequently, the generated 3D bounding boxes often overlap or are overly spaced, preventing realistic object interactions. Based on this, we design a local-global 3D layout generation strategy. First, given a complex text description containing multiple objects, we utilize LLMs to parser the main objects’ words and relative positional relationship. Then we explore the layout planning capability of LLMs to estimate approximate size of each object and return local coordinate. For example, \textit{object 1, [length, width, height], in the left of object 2 [length, width, height] and smaller than object 2}. Subsequently, we place the estimated bounding boxes in global coordinate system according to their relative positional relationships and take into account the perspective prompt in the text to compute their global coordinate. Specific details for each stage will be detailed in the supplementary material.\par
\noindent\textbf{3D Object Reconstruction. }Given a text prompt $y$ and generated 3D layout $B=\left \{ (y_1,b_1),(y_2,b_2),\ldots,(y_n,b_n) \right \}$, where $y_i$ denotes $i$-th object token and $b_i$ is corresponding 3D bounding box. First, we project the 3D bounding box to 2D layout $B^{2D}$ from the front viewpoint, then utilize a layout-guided text-to-image model $G(y,B^{2D})$ to generate the 2D reference image $I_{ref}$. Subsequently, the Segment-Anything Model (SAM) \cite{sam} is utilized to get each instance with broken interaction regions. To avoid inpainting leads to some unreasonable interaction region generation during 3D composition, we directly utilize Large Multi-View Gaussian Model for instance reconstruction without inpainting:
\begin{eqnarray}
I_{ref}= G(y,B^{2D}); I_i=SAM(I_{ref},y_i),
\end{eqnarray}
\begin{eqnarray}
S_{i}=LGM(I_i).
\end{eqnarray}
\subsection{Refinement Stage}
As mentioned in Introduction, existing approaches struggle to generate a compositional 3D scene due to the challenges in synthesizing reasonable interaction region and maintaining multi-view consistency. Based on this, we propose an Inclusive Interactive Collisions strategy (I\textsuperscript{2}C) to ensure that 3D instances are inside the bounding box while allowing Gaussian primitives to collide in the interaction region naturally to achieve reasonable interaction generation. Subsequently, we devise a Multi-View  Adaptive Score Distillation Sampling (MV-ASDS) to distill cross-view prior and layout prior from pre-trained diffusion model by modulating attention map of instance token and spatial token across viewpoints, further achieving multi-view consistency.\par
\noindent\textbf{Inclusive Interactive Collisions. } I\textsuperscript{2}C consists of two constraints: In-Box constraint and Interaction Collision constraint respectively. The former ensures generated compositional 3D assets align with planned 3D layout by judging the position relationship between most Gaussian primitive and its bounding box in a local coordinate system. The latter allows the few Gaussian primitives appearing in reasonable interaction regions, realizing physically plausible and visually coherent object interaction.\par
Firstly, we utilize the K-means algorithm to cluster the global Gaussian primitives $G=\left \{ g_1,g_2,\ldots,g_n \right \}$, and the initial clustering center $C=\left \{ c_1,c_2,\ldots,c_k \right \}$ is the center of the planned instance bounding box $B=\left \{ B_1,B_2,\ldots,B_k \right \}$. Then, the clustering center $C$ is assigned to each Gaussian primitive $g_i$, and the mean of the centers of all the Gaussian primitives in each cluster is computed as the new clustering center. Repeat the above steps until the clustering centers no longer change. The process can be expressed as follows:
\begin{eqnarray}
c_{i}^{t+1} = \frac{1}{\left|G_{i}^{t}\right|} \sum_{g_{j} \in G_{i}^{t}} g_{j},
\end{eqnarray}
where $G_{i}^{t}$ is the set of Gaussian primitives belonging to the clustering center $c_i$ at iteration $t$, and $\left|G_{i}^{t}\right|$ is the number of Gaussian primitives in the set $G_{i}^{t}$.

Subsequently, we constrain most Gaussian primitives to be inside the bounding box of the group which they belong. In addition, we hope the Gaussian primitives to fill the bounding box, avoiding generated 3D assets are too thin and narrow. Therefore, for the set of Gaussian primitives of $i$-th instance $g_{i}=\left \{ g^{1}_{i},g^{2}_{i},\ldots,g^{n}_{i} \right \} $, $B^{max}_{i}$ and $B^{min}_{i}$ are its longest and shortest sides in local coordinate system, respectively, and $M_{i}=\left \{ m^{1}_{i},m^{2}_{i},\ldots,m^{n}_{i}\right \}$ is the local coordinates of all Gaussian primitives of $i$-th instance. Then the In-Box loss can be expressed as follows:
\begin{equation}
\begin{split}
\mathcal{L}_{IB} = \sum_{i=1}^{k} \sum_{j=1}^{N} \Big[ 
    & \mathcal{C}\bigl(\max(m_i^j - B_{i}^{\max}, 0)\bigr)+ \mathcal{C}\bigl(\max(B_{i}^{\min} - m_i^j, 0)\bigr) \Big],
\end{split}
\end{equation}
where $\mathcal{C}$ is the clamp function to ensure that the difference value is non-negative. Minimizing the above In-Box loss until the proportion of the number of Gaussian primitives in the group is greater than threshold and stop the calculation.\par
The remaining Gaussian primitives which are on the outside of the bounding box and in the space of neighboring bounding boxes, are used to generate interaction regions. As shown in Fig. \ref{fig:short-b}, interaction collision tend to occur in the region near the midpoint of the line connecting the centers of the neighboring bounding boxes. Therefore, we design Interaction Collision constraints to guide the remaining Gaussian primitives toward the interaction region for collision and interaction. Based on triangle trilateral relationship theory in geometry, i.e., the sum of two sides of a triangle is greater than or equal to the third side, for the $i$-th set of Gaussian primitives beyond the bounding box, we compute the interaction collision loss with the neighboring $j$-th bounding box:
\begin{eqnarray}
\mathcal{L_{IC}}=\sum_{k=1}^{N} max(0,(d_{k,i}+d_{k,j})-d_{i,j}-\tau ),
\end{eqnarray}
where $d_{i,j}$ is the European distance between the $i$-th and $j$-th bounding boxes, $d_{k,i}$ and $d_{k,j}$ denote the distance from the $k$-th Gaussian primitive to the $i$-th bounding box and the $j$-th bounding box. $\tau$ is a hyperparameter to avoid too many Gaussian primitives densely clustered in a small region, which is set to 5\% of the maximum bounding-box diagonal.\par

The total loss in I\textsuperscript{2}C can be expressed as follows:
\begin{eqnarray}
\mathcal{L}_{{\mathcal{I}^2}\mathcal{C}}=\lambda_{\mathcal{{IB}} } \mathcal{L_{IB}}+\lambda_{\mathcal{{IC}} }\mathcal{L_{IC}},
\end{eqnarray}
where $\lambda_{\mathcal{{IB}}}$ and $\lambda_{\mathcal{{IC}}}$ are two hyperparameters.\par
\noindent\textbf{Multi-View Adaptive Score Distillation Sampling. }
Traditional SDS cannot accurately achieve positional adjustment and spatial optimization. Inspired by \cite{chen2024comboverse,zhou2024layout}, we enhance the attention scores corresponding to the tokens describing spatial relations to improve spatial expressions in diffusion process. However, only enhancing the attention of spatial tokens while ignoring the instance tokens leads to unclear textures and geometric inconsistency. To solve this issue, we propose Adaptive Score Distillation Sampling (ASDS) to not only enhance the attention of spatial token but also modulate the cross attention map of instance token according to the segmentation region, which can be represented as:
\begin{eqnarray}
{A_{i}} =  softmax(\frac{Q_{i}K_{i}^T}{\sqrt{d} } )  ,
\end{eqnarray}
\begin{eqnarray}
A_{i}': = \left\{\begin{array}{ll}
k \cdot A_{i}, & \text { if } \quad i  =  i_{spa} \\
A_i \cdot M_i +\frac{A_i}{k},  \cdot(1-M_i), & \text { if } \quad i  =  i_{ins}\\
A_{i}, & \text { otherwise } . 
\end{array}\right.,
\end{eqnarray}
where $A_i$ and $A_{i}^{'}$ individually represent the original and modulated cross-attention maps corresponding to the $i$-th token. $i_{spa}$ and $i_{ins}$ are spatial token and instance token respectively. While $M_i$ indicates the mask region of $i$-th token, and $k$ is a constant, which is set to 1.25 in experiment. 

Sole ASDS distills appearance feature and spatial layout information from the pre-trained text-to-image diffusion model, which cannot guarantee the consistency of multiple viewpoints, leading to the Janus problem. To alleviate this issue, we introduce Multi-View Adaptive Score Distillation Sampling (MV-ASDS) to distill cross-view prior from the multi-view diffusion model, which ensures the multi-view consistency of the generated compositional 3D scene. Unlike previous methods, we apply MV-ASDS not only to the compositional 3D scene but also to individual entity to further enhance the multi-view consistency. Similar to SDS, we render an image from the 3D representation at viewpoint $v$, then add random Gaussian noise to it. Subsequently, utilize Stable Zero123 $\epsilon_{\phi^{*}}$ to predict noise and denoise. Then minimize the predicted noise $\epsilon_{\phi^{*}}(x_{t};y,t,v)$ with the added noise to align the rendered image and the reference image at viewpoint $v$. The process can be expressed as follows:
\begin{multline}
\nabla_{\theta}\mathcal{L}_{\text{MV-ASDS}}(\epsilon_{\phi^{*}},\theta,v,y)= \mathbb{E}_{t,\epsilon,v}\Bigl[
   w(t)\bigl(\epsilon_{\phi^{*}}(x_{t};y,t,v)-\epsilon\bigr)
   \frac{\partial g(\theta,v)}{\partial\theta}
\Bigr],
\end{multline}
where $\epsilon _{\phi^{*}}(x_{t};y,t,v)$ is the predicted noise calculated with the modulated attention maps which focus on the spatial words and instance tokens.\par
When applying MV-ASDS to every entity, we scale the 3D bounding box of an entity Gaussian $\theta _i$ with a global volumetric space $B_{global}$. Then we compute the shift parameter $\beta$ and scale parameters $\gamma$ to realize the transformation for the center position:
\begin{eqnarray}
\beta  = Mean(B_i);
\gamma  =  B_{global}/B_i .
\end{eqnarray}

\section{Experiments}

\subsection{Experiment Setting}
\textbf{Implementation details. }Our I\textsuperscript{2}C-3D is implemented by ThreeStudio \cite{threestudio}. We leverage GLIGEN \cite{gligen} to generate reference image and pre-trained LGM \cite{lgm} is used to generate individual 3D object. Zero-123 is used to realize MV-ASDS and the guidance timestep is set to 50. We optimize Gaussian primitives through $\mathcal{L}_{{\mathcal{I}^2}\mathcal{C}}$ for 1000 iteration and the hyperparameter $\lambda_{\mathcal{{IB}} }$ and $\lambda_{\mathcal{{IC}} }$ are set to 0.15 and 1.0 respectively. Moreover, the threshold in In-Box Loss is set to 0.95.\par
\noindent\textbf{Dataset. }For a fair comparison, we utilize the dataset provided by ComboVerse \cite{comboverse} for evaluation, which consists of 50 textual descriptions, 50 corresponding real images, and 50 images generated by Stable Diffusion. We utilize the textual data to evaluate the performance in text-to-3D and use real images to evaluate the generation capability in single image-to-3D generation.\par
\noindent\textbf{Baselines. }To enable a more comprehensive evaluation, we conduct experiments on three tasks: text-to-3D, single image-to-3D and compositional 3D generation. For text-to-3D, we compare our method with representative approaches based on NeRF and 3DGS. Specifically, DreamFusion \cite{poole2022dreamfusion}, Latent-NeRF \cite{lnerf}, sjc \cite{sjc} and Magic3D \cite{magic3d} adopt NeRF as the 3D representation, while GaussianDreamer \cite{gaussiandreamer}, gsgen \cite{gsgen}, and GraphDreamer \cite{graphdreamer} are built upon 3DGS. For single-image-to-3D, we select four competitive baselines for comparison, including InstantMesh \cite{instantmesh}, Hunyuan3D \cite{hunyuan3d}, ReconViaGen \cite{chang2025reconviagen}, and TRELLIS \cite{trellis}. In addition, we compare with two compositional 3D generation methods: Layout-your-3D \cite{layout-your-3d} and ComboVerse \cite{comboverse}. More comparisons with the 3D scene generation method (MIDI \cite{huang2025midi}) are provided in the \textbf{Supplementary Materials}.\par
\noindent\textbf{Evaluation metrics. }Following common practice \cite{reparo}, we choose CLIP-Score \cite{clip} to evaluate the semantic alignment between generated 3D assets and text. However, CLIP-Score struggles to assess the geometry consistency and spatial relationships. Therefore, following by \cite{gpt-4v,sds}, we utilize GPT-4V to provide a more comprehensive assessment of 3D assets from four aspects: Prompt Alignment, Spatial Arrangement, Geometric Fidelity and Scene Quality. Moreover, for single image-to-3D, we leverage CLIP-Sim, PSNR, SSIM \cite{ssim} and LPIPS \cite{lpips} to evaluate the similarity of the rendered image with the reference image and generation quality of 3D asset. Additionally, we also calculate the training time as an efficiency metric.\par


\subsection{Comparisons with Existing Methods}
\textbf{Quantitative results. }Table \ref{table1} shows quantitative comparison results between our I\textsuperscript{2}C-3D and various SOTA models on the benchmark, which demonstrates our method significantly outperforms previous approaches in Semantic Alignment, Spatial Arrangement, Geometric Consistency and Scene Quality. As for training time, compositional 3D generation needs longer time to optimize all objects, which is related with the number of objects. Despite a slightly longer training time than Latent-NeRF and GaussianDreamer, both qualitative and quantitative results indicate that our I\textsuperscript{2}C-3D improve generation quality by a large margin. In addition, we provide more quantitative experiments compared with single image-to-3D methods in \textbf{Supplementary Materials}, which indicates that our proposed I\textsuperscript{2}C-3D performs well in CLIP-Sim and PSNR, and achieves competitive results in SSIM and LPIPS.
\begin{figure*}[t]
\centering
\includegraphics[width=1.0\columnwidth]{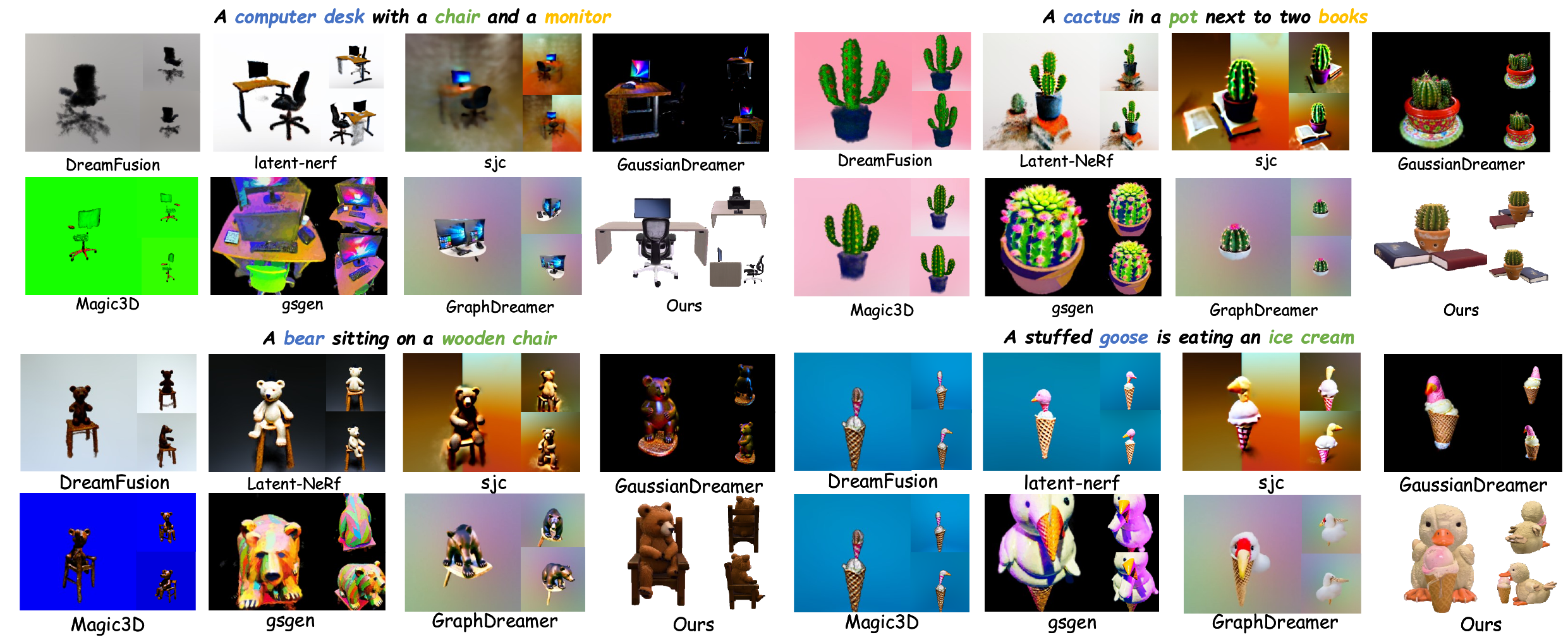}
\caption{Qualitative comparison between our approach and previous methods.}
\label{fig4}
\end{figure*}
\renewcommand{\arraystretch}{1.3}
\begin{table*}[t]
\centering
\caption{\textbf{Quantitative comparisons of I\textsuperscript{2}C-3D and other competing text-to-3d methods on ComboVerse benchmark.} The highest and second scores in each section of the table are highlighted in \textbf{bold} and \underline{underlined} respectively.}
\resizebox{\textwidth}{!}{
\begin{tabular}{cccccccc}
\hline
\multirow{2}{*}{Method} & \multirow{2}{*}{CLIP-Score $\uparrow$} & \multirow{2}{*}{3D} & \multicolumn{4}{c}{GPT-4V}                                                  & \multirow{2}{*}{Time $\downarrow$} \\ \cline{4-7}
                        &                             &                     & Prompt Alignment $\uparrow$ & Spatial Arrangement $\uparrow$ & Geometric Fidelity $\uparrow$ & Scene Quality $\uparrow$ &                       \\ \hline
DreamFusion             & 0.273                       & NeRF                & 50.6             & 51.3                & 50.9               & 28.5          & 6hours                \\
Latent-NeRF             & 0.299                       & NeRF                & 67.9             & 70.6                & 68.4               & 67.7          & \textbf{15min}        \\
sjc                     & 0.303                       & NeRF                & 65.2             & 69.7                & 67.3               & 71.2          & 2hours                \\
Magic3D                 & 0.276                       & NeRF                & 55.1             & 53.4                & 54.5               & 52.4          & 5.3hours              \\ \hline
GaussianDreamer         & 0.295                       & 3DGS        & 78.4             & 75.3                & 74.3               & 67.5          & \textbf{15min}        \\
gsgen                   & \underline{0.304}                 & 3DGS        & \underline{80.9}       & 72.9                & 73.7               & 70.2          & 3hours                \\
GraphDreamer            & 0.286                       & 3DGS        & 78.5             & \underline{80.3}          & \underline{81.1}         & \underline{79.6}    & 3hours                \\
\textbf{I\textsuperscript{2}C-3D (Ours)}                    & \textbf{0.314}              & 3DGS        & \textbf{92.7}    & \textbf{84.5}       & \textbf{82.6}      & \textbf{85.3} & \underline{28min}           \\ \hline
\end{tabular}
}

\label{table1}
\end{table*}

\noindent\textbf{Qualitative results. }As shown in Fig. \ref{fig4}, our I\textsuperscript{2}C-3D can generate compositional 3D assets that not only enable reasonable interaction regions, but also maintain multi-view consistency. However, other methods struggle to generate interaction regions and suffer from the Janus Problem. For example, DreamFusion \cite{poole2022dreamfusion} fails to generate multiple objects based on given complex text, mainly due to the difficulty of SDS in optimizing multiple objects. While GaussianDreamer \cite{gaussiandreamer}, Magic3D \cite{magic3d} and sjc \cite{sjc} can improve the appearance of the 3D assets but only generate partial objects of given text. In addition, gsgen \cite{gsgen} and GraphDreamer \cite{gaussiandreamer} suffer from multi-view inconsistency and unreasonable interaction regions. Moreover, Latent-NeRF \cite{lnerf} and sjc \cite{sjc} can generate good geometric structures by introducing the mesh prior, but produce artifacts and messy spatial relationships. In contrast, I\textsuperscript{2}C-3D can generate high-quality geometry and texture for each object while preserving good spatial relationships.\par

\begin{figure*}[t]
\centering
\includegraphics[width=1.0\columnwidth]{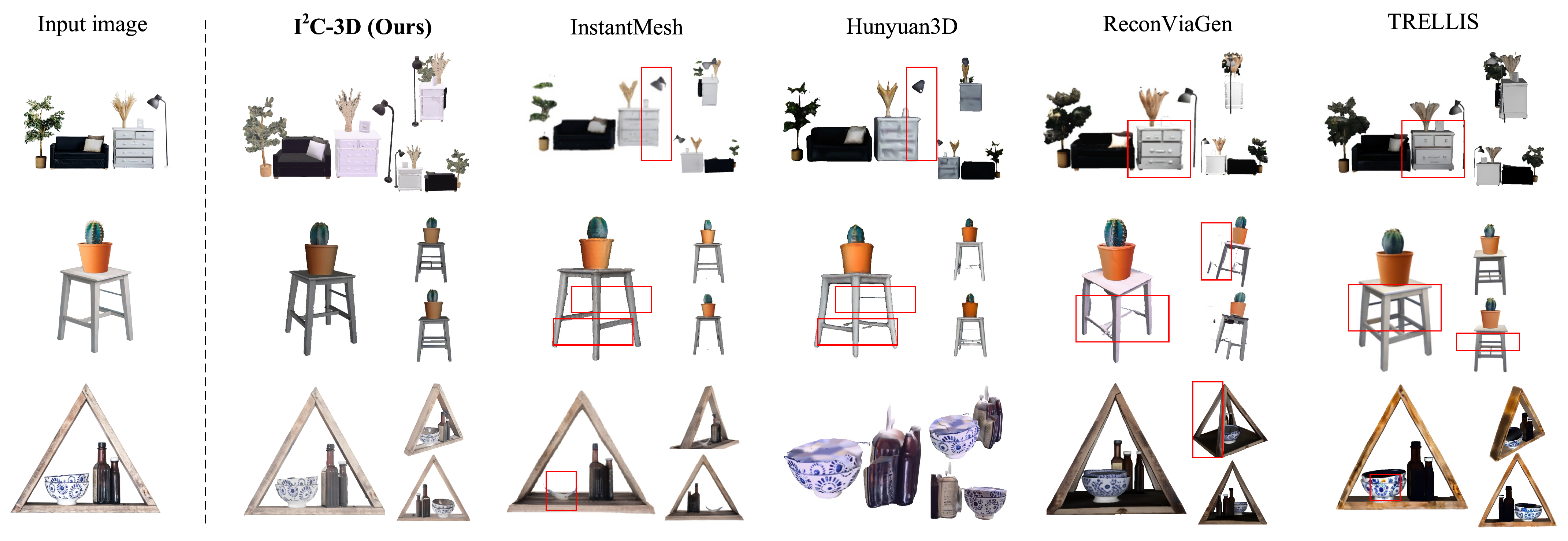}
\caption{Qualitative comparisons between our I\textsuperscript{2}C-3D and prominent methods for single image-to-3D generation.}
\label{fig6}
\end{figure*}

In single image-to-3D tasks, other methods suffer from object incompleteness and cross-view inconsistency (shown in Fig. \ref{fig6}). For instance, in the first example, InstantMesh and Hunyuan3D generate partially missing light structures, while cabinets generated by ReconViaGen and TRELLIS are both unfaithful. In the last example, ReconViaGen emerges severe multi-view inconsistency, InstantMesh and Hunyuan3D fail to reconstruct the entire scene, and the appearance of the bowl generated by TRELLIS is inaccurate. On the contrary, our method reconstructs geometrically consistent 3D assets with reasonable interaction, faithfully preserving object appearance as well as the spatial relationships depicted in the input image.\par

\begin{figure*}[htbp]
\centering
\includegraphics[width=1.0\columnwidth]{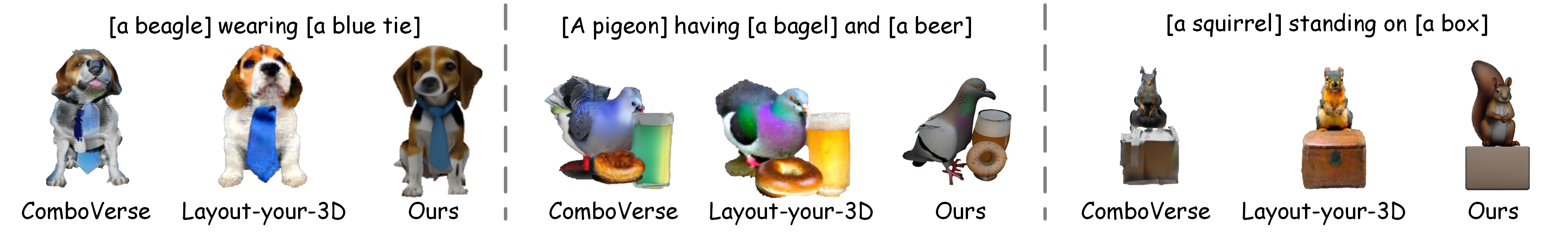}
\caption{Comparisons between our I\textsuperscript{2}C-3D and compositional 3D generation methods.}
\label{fig7}
\end{figure*}
\begin{figure*}[htbp]
\centering
\includegraphics[width=1.0\columnwidth]{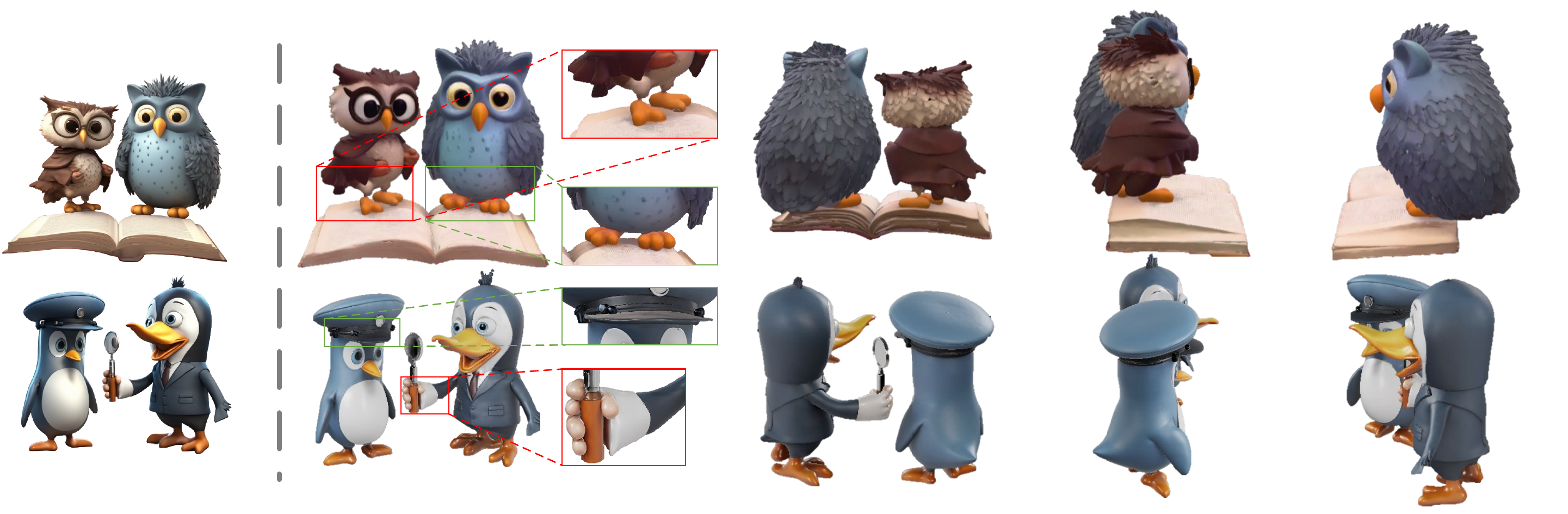}
\caption{Challenging cases that contain complex interactions.}
\label{multi}
\end{figure*}

Moreover, we compared our method with open-source compositional 3D generation methods. As displayed in Fig. \ref{fig7}, the interaction area in our generated 3D assets is more reasonable. For instance, the wearing relationship between the tie and the beggle is expressed more naturally. Meanwhile, we also present some challenging interactive cases in Fig. \ref{multi} (e.g., more humanized interactions), which demonstrates that our method achieves competitive results in complex interaction tasks. What's more, we provide more visual comparisons in the \textbf{supplementary material} and support video material to display our generation results, you can click to watch the results more intuitively.

\vspace{-0.5cm}


\vspace{-5pt}
\begin{wrapfigure}{r}{0.45\columnwidth}
\vspace{-15pt}
\centering
\includegraphics[width=\linewidth]{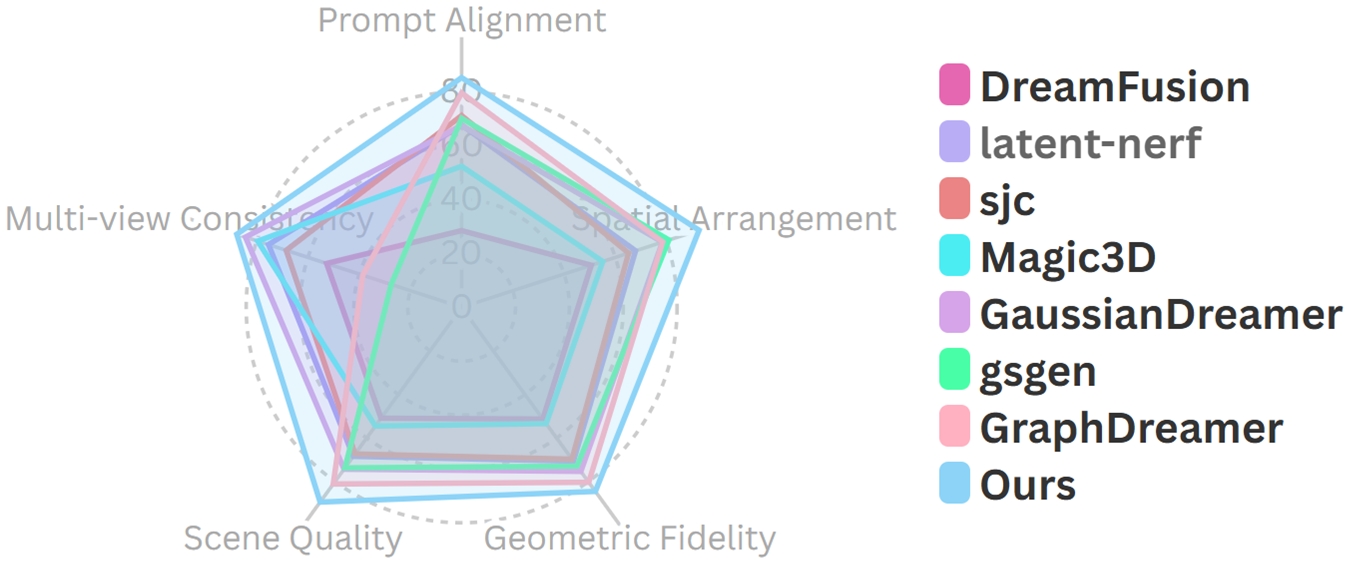}
\caption{Human preference results.}
\label{user}
\vspace{-30pt}
\end{wrapfigure}

\noindent\textbf{User Study. }We conducte a questionnaire to explore users' preference through scoring our method and other baselines across 20 prompts including more than 50 objects in terms of Prompt Alignment, Spatial Arrangement, Geometric Fidelity, Scene Quality and Multi-View Consistency. The visualization of human preference are shown in Fig. \ref{user}, which shows that our model significantly outperforms other methods in above five aspects.\par

\subsection{Extended Applications: 3D Editing}
\begin{figure}[htbp]
\centering
\includegraphics[width=1.0\columnwidth]{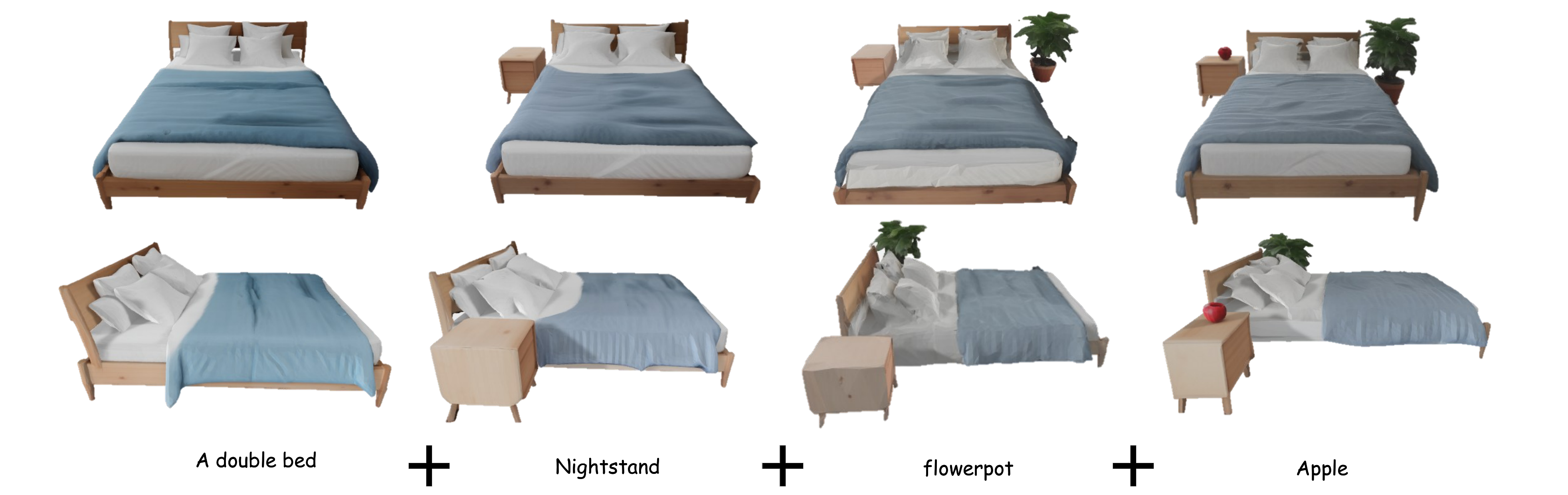}
\caption{Progressive 3D Editing examples of our I\textsuperscript{2}C-3D.}
\label{fig8}
\end{figure}

As illustrated in Fig.~\ref{fig8}, our approach supports progressive 3D editing for compositional scenes by incrementally inserting new objects according to textual instructions. Starting from a single object prompt (a double bed), our I\textsuperscript{2}C-3D first reconstructs a stable 3D asset. Subsequently, additional objects (nightstand, flowerpot and apple) can be inserted step by step. Importantly, each newly added object is integrated while preserving the geometric structure, spatial relationships and interaction consistency of the existing scene. Moreover, the previous objects remain stable without distortion or displacement and the inserted objects are placed in physically plausible positions with coherent cross-view appearance. This progressive compositional editing capability demonstrates that our method is not limited to one-shot generation, but can flexibly support iterative scene refinement and interactive 3D asset generation.

\subsection{Ablation Study}
\begin{figure}[htbp]
\centering

\begin{minipage}[t]{0.5\linewidth}
\vspace{0pt}   
\captionsetup{type=table}
\setlength{\abovecaptionskip}{2pt}
\setlength{\belowcaptionskip}{2pt}
\caption{Quantitative ablation experiment result on key components.}
\vspace{0.3em}
\renewcommand{\arraystretch}{1.6}
\resizebox{\linewidth}{!}{
\begin{tabular}{ccccc}
\hline
Component & Full Setting & w/o MV-ASDS & w/o Adaptive & w/o I\textsuperscript{2}C \\ \hline
CLIP-Score $\uparrow$ & \textbf{0.314} & 0.275 & 0.293 & \underline{0.307} \\
Prompt Alignment $\uparrow$ & \textbf{92.7} & 40.3 & \underline{90.5} & 84.2 \\
Spatial Arrangement $\uparrow$ & \textbf{84.5} & 72.8 & \underline{75.4} & 43.3 \\
Geometric Fidelity $\uparrow$ & \textbf{82.6} & 45.9 & \underline{74.3} & 72.8 \\
Scene Quality $\uparrow$ & \textbf{85.3} & 47.2 & \underline{70.8} & 67.4 \\ \hline
\end{tabular}}
\label{table2}
\end{minipage}
\hfill
\begin{minipage}[t]{0.48\linewidth}
\vspace{0pt}   
\centering
\includegraphics[width=\linewidth]{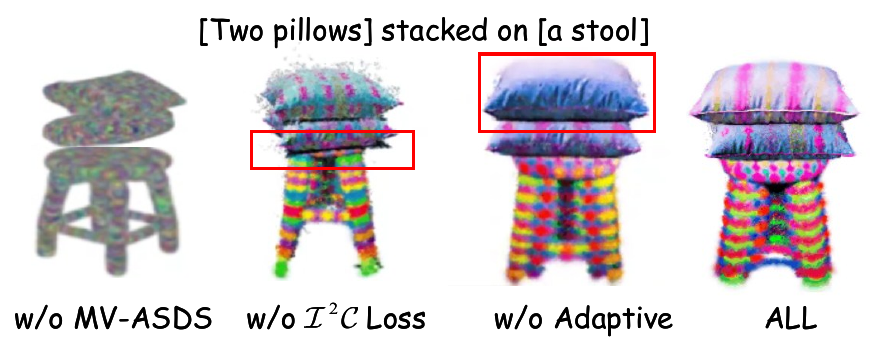}
\vspace{-0.5em}
\captionsetup{type=figure}
\setlength{\abovecaptionskip}{2pt}
\setlength{\belowcaptionskip}{0pt}
\caption{Visual ablation results on key components.}
\label{fig9}
\end{minipage}

\end{figure}

\textbf{Effectiveness of Multi-View Adaptive Score Distillation Sampling. }Results shown in Table \ref{table2} and Fig. \ref{fig9} indicate the poor multi-view consistency and the disappeared texture appearance when discarding the MV-ASDS. For example, a pillow with a round front becomes square on the sides and the texture appearance of objects is missing.\par
\noindent\textbf{Effectiveness of Inclusive Interactive Collisions. } Without I\textsuperscript{2}C, the generated 3D asset is not full and will have many outlier Gaussian primitives. In addition, the interaction regions are expressed ambiguously. As shown in column 2 of Fig. \ref{fig9}, there are black artifacts in the interaction region between the pillow and the stool.\par
\noindent\textbf{Effectiveness of Adaptive Score Distillation Sampling. }It can also be observed in the column 3 of Fig. \ref{fig9}, without ASDS, the generated 3D asset lacks texture details. For instance, the floral detail on the pillows is missing and the texture of the stool is blurred. Additionally, as illustrated in Table \ref{table2}, ASDS promotes prompt alignment, geometric fidelity and scene quality significantly.\par

\begin{figure}[htbp]
\centering

\begin{minipage}[t]{0.5\linewidth}
\vspace{0pt}\strut
\captionsetup{type=table}
\setlength{\abovecaptionskip}{2pt}
\setlength{\belowcaptionskip}{2pt}
\caption{Ablation on $\tau$.}
\renewcommand{\arraystretch}{1.2}
\resizebox{\linewidth}{!}{
\begin{tabular}{lccc}
\hline
Ablation & $\tau=0.02$ & $\tau=0.2$ & $\tau=0.5$ \\ \hline
CLIP-Score $\uparrow$ & 0.310 & \textbf{0.314} & \underline{0.312} \\
Prompt Alignment $\uparrow$ & 91.8 & \textbf{92.7} & \underline{92.3} \\
Spatial Arrangement $\uparrow$ & 70.7 & \textbf{84.5} & \underline{72.6} \\
Scene Quality $\uparrow$ & 82.1 & \textbf{85.3} & \underline{83.4} \\ \hline
\end{tabular}}
\label{table_tau}
\end{minipage}
\hfill
\begin{minipage}[t]{0.4\linewidth}
\vspace{0pt}\strut
\captionsetup{type=figure}
\setlength{\abovecaptionskip}{2pt}
\setlength{\belowcaptionskip}{2pt}
\centering
\includegraphics[width=\linewidth]{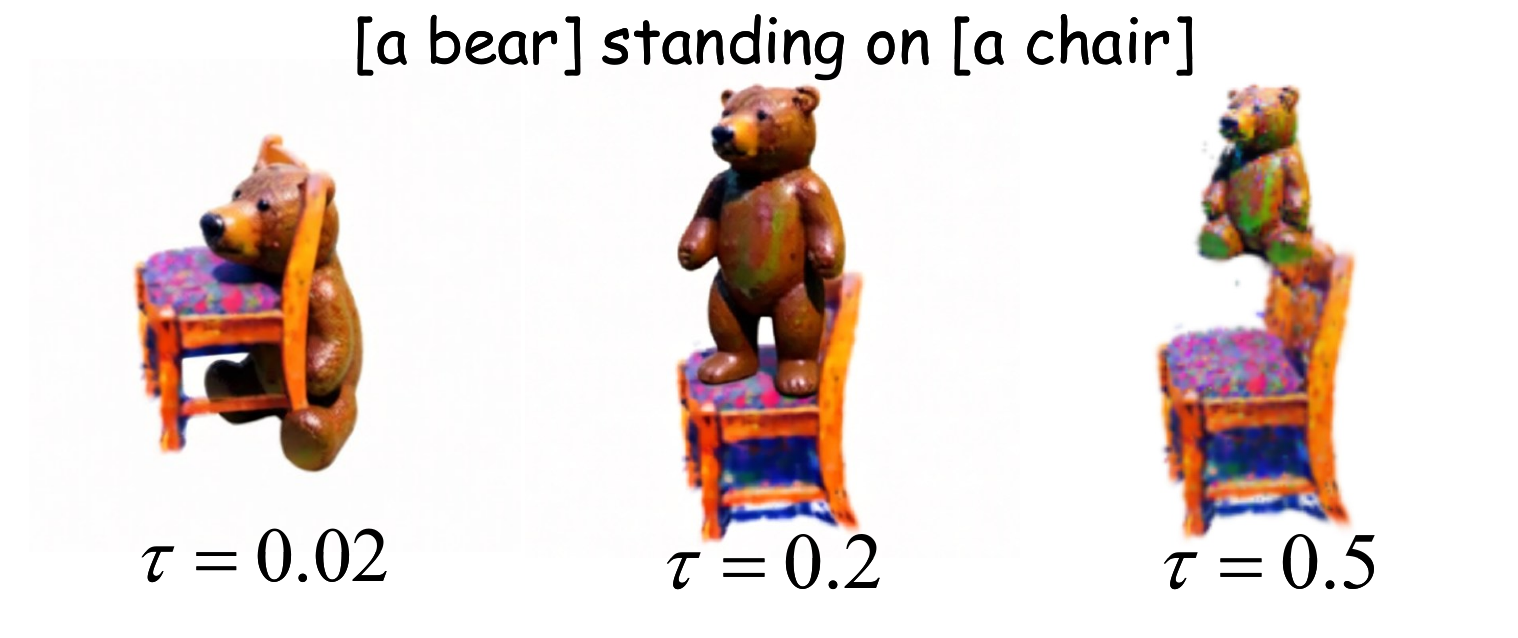}
\caption{Ablation on $\tau$.}
\label{fig_tau}
\end{minipage}

\end{figure}

\noindent\textbf{Hyperparameter Ablation on $\tau$.}
We conduct an ablation study on the interaction margin hyperparameter $\tau$, which controls the minimum separation distance between object bounding boxes during optimization. In our implementation, $\tau$ is set to 5\% of the maximum diagonal of bounding box, which provides a scale-adaptive margin across different scenes. Table \ref{table_tau} reports the quantitative evaluation and we observe that moderate values achieve better spatial arrangement and scene quality, while extremely small or large values degrade interaction realism. Additionally, the visual ablation result is shown in Fig. \ref{fig_tau}. When $\tau$ is too small ($\tau=0.02$), the margin constraint becomes insufficient, leading to noticeable object penetration and unrealistic physical contact. When $\tau$ is too large ($\tau=0.5$), the enforced separation becomes overly strong, causing visible gaps between interacting objects. In contrast, a moderate value ($\tau=0.2$) yields stable contact relationships and physically plausible interactions. Overall, the above results demonstrate that setting $\tau$ to 5\% of the maximum bounding-box diagonal achieves a good balance between preventing penetration and avoiding excessive separation, leading to more realistic interaction modeling.

\vspace{-0.5cm}

\subsection{More experiments in challenging scenarios}
\begin{figure*}[htpb]
  \centering
   \includegraphics[width=1.0\linewidth]{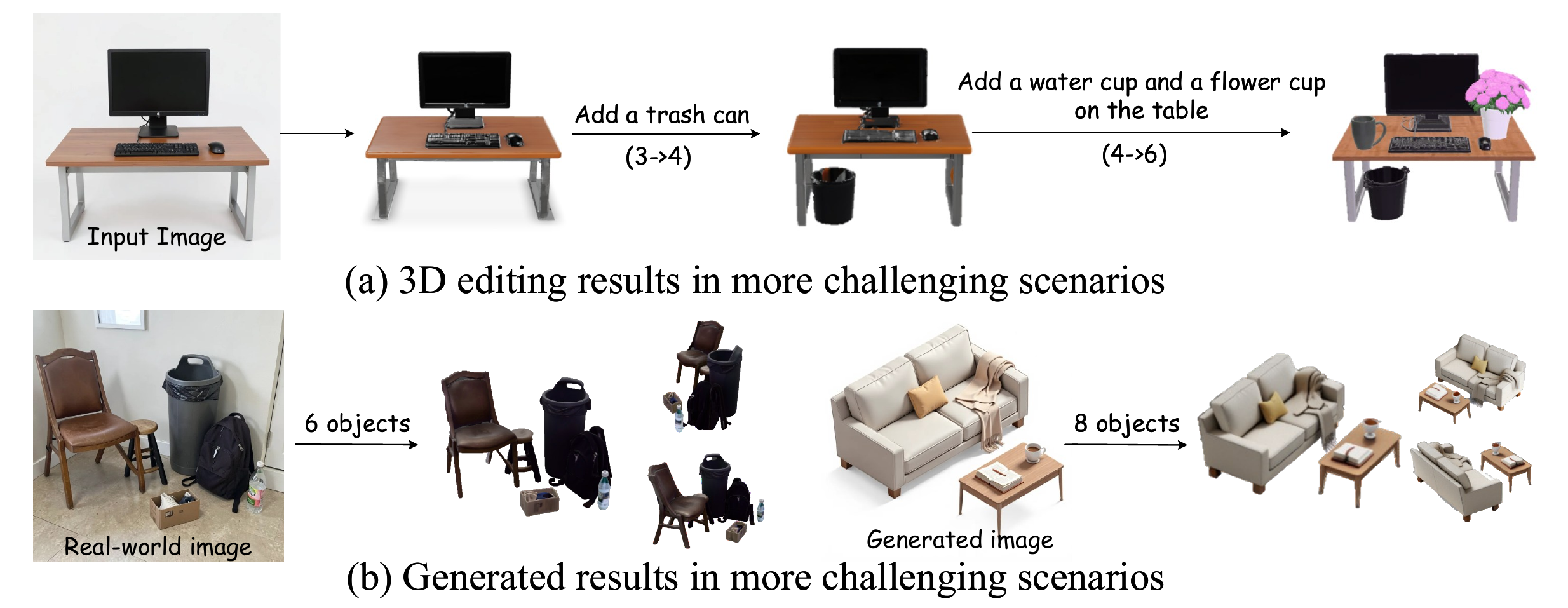}
   \caption{More experiments in challenging scenarios.}
   \label{more}
\end{figure*}

\vspace{-0.2cm}
We conduct more experiments to further evaluate the robustness of our I\textsuperscript{2}C-3D under more challenging scenarios. As illustrated in Fig.~\ref{more}(a), we progressively increase scene complexity by inserting additional objects. Starting from a input with 3 object, our I\textsuperscript{2}C-3D successfully extends the scene to 4 and then 6 objects while preserving spatial relationships and interaction coherence. Despite the increased object count and potential occlusions, the generated 3D assets remain geometrically stable and semantically consistent with the input instructions. What’s more, the generation results under real-world and generated scenarios are shown in Fig.~\ref{more}(b), which further demonstrates our I\textsuperscript{2}C-3D maintains stable interaction quality and cross-view consistency in challenging scenarios.

\section{Conclusion}

In this paper, we propose a novel framework for multi-view consistent compositional 3D generation, termed I\textsuperscript{2}C-3D. Unlike existing methods that suffer from unreasonable interaction generation, our I\textsuperscript{2}C-3D can generate physically plausible and visually coherent interaction region. Specifically, we first estimate the 3D layout based on the given text and generate a reference image for single-object reconstruction. Subsequently, we propose an Inclusive Interactive Collisions strategy to constrain Gaussian primitives in the interaction region appearing at reasonable interaction locations naturally for interaction region generation. Additionally, Multi-View Adaptive Score Distillation Sampling is designed to distill multi-view consistency prior and layout prior from pre-trained diffusion model by adaptive attention modulation across viewpoints, further achieving multi-view consistency. Extensive experiments indicate that our I\textsuperscript{2}C-3D outperforms existing methods in terms of multi-view consistency and rational interaction region generation.\par


%
%
\bibliographystyle{splncs04}
\bibliography{main}

\end{document}